\documentclass{article}

\usepackage{general_template}

\usepackage[utf8]{inputenc}
\usepackage[T1]{fontenc}
\usepackage{lmodern}
\usepackage{amsmath}
\usepackage{amsfonts}
\usepackage{amsthm}
\usepackage{array}
\usepackage{booktabs}
\usepackage{colortbl}
\usepackage{float}
\usepackage{graphicx}
\usepackage{placeins}
\usepackage{makecell}
\usepackage{microtype}
\usepackage{multirow}
\usepackage{nicefrac}
\usepackage{pifont}
\usepackage[most]{tcolorbox}
\usepackage{xspace}
\usepackage{enumitem}
\usepackage{wrapfig}
\usepackage{arydshln}
\usepackage[ruled,vlined]{algorithm2e}

\definecolor{chronoblue}{HTML}{1B6A8F}
\definecolor{chronoyellow}{HTML}{FFF8D8}
\definecolor{chronoframe}{HTML}{676A42}
\definecolor{discussionback}{HTML}{EEF7EE}
\definecolor{discussionframe}{HTML}{333A33}

\usepackage[
  colorlinks=true,
  linkcolor=chronoblue,
  citecolor=chronoblue,
  urlcolor=magenta
]{hyperref}

\newtheorem{theorem}{Theorem}
\newtheorem{property}[theorem]{Property}
\newtheorem{proposition}[theorem]{Proposition}

\definecolor{lightgrayrow}{RGB}{245,245,245}
\definecolor{lightgrayhead}{RGB}{228,228,228}
\definecolor{commentcolor}{RGB}{70,120,70}
\definecolor{headergray}{RGB}{224,224,224}
\definecolor{rowgray}{RGB}{242,242,242}
\definecolor{oursblue}{RGB}{220,240,245}
\arrayrulecolor{black}
\newcommand{\mycomment}[1]{\textcolor{commentcolor}{\# #1}}

\newcommand{\ChronoLock}{{\fontfamily{lmtt}\selectfont \textbf{ChronoLock}}\xspace}

\definecolor{blackish}{rgb}{0.2, 0.2, 0.2}
\definecolor{grayish}{rgb}{0.95, 0.95, 0.95}

\newcommand{\answerTODO}[1][]{\textcolor{red}{\bf [TODO]}}

\newtcolorbox{chronobox}{
  colframe=black!65,
  colback=yellow!5,
  boxrule=0.9pt,
  arc=4mm,
  left=10pt,
  right=10pt,
  top=7pt,
  bottom=7pt,
  before skip=4pt,
  after skip=4pt,
  boxsep=0pt
}
\newtcolorbox{discussionbox}{
  colframe=discussionframe,
  colback=discussionback,
  boxrule=1.0pt,
  arc=3mm,
  left=7pt,
  right=7pt,
  top=3pt,
  bottom=3pt,
  before skip=4pt,
  after skip=4pt,
  boxsep=0pt
}

\definecolor{templateabstractback}{RGB}{247,247,247}
\newenvironment{templateabstract}{
  \begin{tcolorbox}[
    colback=templateabstractback,
    colframe=white,
    boxrule=0pt,
    arc=10pt,
    left=16pt,
    right=16pt,
    top=12pt,
    bottom=12pt,
    width=\textwidth,
    before skip=10pt,
    after skip=10pt
  ]
}{
  \end{tcolorbox}
}

\begin{document}

\icmldate{\today}
\icmltitlerunning{\ChronoLock}
\icmltitle{\ChronoLock: Protecting Videos from Unauthorized Text-to-Video Personalization}

\begin{icmlauthorlist}
Jiaming He$^{1\,}$, Jiashu Zhang$^{2\,}$, Guanyu Hou$^{3\,}$, Shuhan Ye$^{1\,}$, Hanwei Zhu$^{1\,}$, Yi Yu$^{2\,}$, \\ Xudong Jiang$^{1\,}$
\end{icmlauthorlist}

$^{1\,}$Nanyang Technological University  \quad $^{2\,}$Jilin University \quad $^{3\,}$The University of Manchester 

\begin{templateabstract}
\textbf{Abstract.}
Text-to-video (T2V) diffusion models have made it increasingly easy to synthesize realistic and temporally coherent videos, while recent personalization techniques allow such models to imitate a specific subject, style, or motion pattern from only a few reference clips. This capability creates a new data-misuse risk: videos shared online can be collected and used for unauthorized T2V fine-tuning. 
Existing protective perturbations are mainly designed for image recognition or text-to-image personalization, and therefore focus on corrupting static appearance cues rather than the temporal denoising dynamics that make video personalization possible. 
To address this gap, we introduce \ChronoLock, the first proactive protection framework that makes released videos difficult to exploit for unauthorized T2V personalization. 
\ChronoLock targets the motion-learning process directly by optimizing bounded perturbations over temporal denoising trajectories. 
It first disrupts intra-chunk temporal adaptation with a diffusion objective that combines fitting error, frame-relative denoising relations, and adjacent-frame variation, and then enlarges inter-chunk boundary mismatch to weaken long-range motion continuity. 
Transformation-sampled updates further improve robustness to common preprocessing operations. 
Experiments on UCF Sports and HMDB51 with popular T2V backbones and personalization scheme show that \ChronoLock consistently reduces motion imitation under automatic metrics and human evaluation.
Code is available at \href{https://github.com/burst8797/ChronoLock}{Github}. 

\end{templateabstract}

\section{Introduction}
\vspace{-2mm}
Text-to-video (T2V) diffusion models are rapidly changing how dynamic visual content is produced.
Recent systems can synthesize realistic and temporally coherent videos from natural language
prompts \citep{ho2022imagenvideo,singer2023makeavideo,villegas2023phenaki,chen2023videocrafter1}.
Meanwhile, emerging personalization paradigms allow a pre-trained generator to adapt to a specific
subject, style, o and motion pattern using only a few reference examples
\citep{wu2023tuneavideo,guo2024animatediff,zhao2024motiondirector}. While such capability benefits
creative editing, digital avatars, and customized media production, it also raises a direct copyright
and privacy risk: \textit{videos shared online may be collected and used to fine-tune a T2V model
without the owner's consent}.

\textbf{From Static Cloaks To Temporal Cloaks.}
The risk has been studied earlier in text-to-image (T2I) personalization, where methods such as
Textual Inversion~\citep{gal2022textualinversion} and DreamBooth~\citep{ruiz2023dreambooth} learn a personal concept from a small reference set
. Prior defenses show that user data can be
proactively protected by imperceptible perturbations before release, including unlearnable examples
\citep{huang2021unlearnable, li2025survey, li2025detecting, zhu2026unlearnable, wang2024unlearnable,xu2025not}, face-recognition cloaking
\citep{shan2020fawkes,cherepanova2021lowkey}, and recent defenses against unauthorized T2I
customization or style imitation
\citep{shan2023glaze,le2023antidreambooth,shan2024nightshade,liu2024disrupting,liu2024metacloak}.
These methods mainly corrupt appearance learning, pixel-level reconstruction, or token-concept
association. However, a video is not merely a stack of static images. For a video, the T2V personalization paradigm
must also learn the continuous semantics in the additional temporal dimension through temporal attention, temporal transformer blocks and motion
modules. Recent studies~\citep{yuan2025videorefer, wang2026timerefine, xie2025star, schiber2025tempocontrol, he2026tear}
also suggest that video-specific failures can arise from temporal sequencing rather than from any
single frame. Hence, a research question arises naturally:

\begin{chronobox}
\vspace{-1mm}
\small
\textit{How can we protect released videos by corrupting the temporal denoising evidence used for
motion personalization, rather than only hiding static semantic?} 
\vspace{-1mm}
\end{chronobox}

\begin{figure*}
\centering
\includegraphics[width=\textwidth]{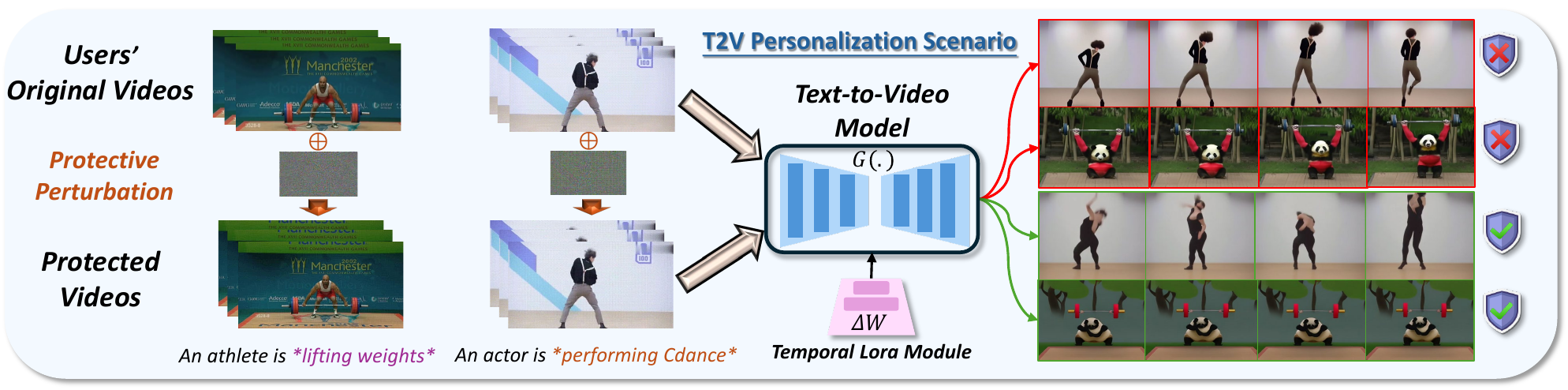} 
\vspace{-6mm}
\caption{\textbf{Scenario of unauthorized T2V personalization.}}
\label{fig:teaser}
\end{figure*}

\textbf{The Present Framework: \ChronoLock.}
In response to this question, we propose \ChronoLock, a proactive protection framework for
preventing videos from unauthorized T2V personalization. The application scenario of \ChronoLock is illustrated in Fig.~\ref{fig:teaser}. Our key observation is that unauthorized
T2V fine-tuning depends on learnable temporal denoising trajectories. If bounded perturbations
corrupt these trajectories, the adapted temporal module can be driven toward unstable motion cues
even when the released video remains visually close to the original. Accordingly, \ChronoLock
partitions a video into contiguous chunks, enlarges temporal fitting errors inside each chunk, and
then disrupts boundary-level denoising continuity that supports coherent long-range generation.

\textbf{Contributions.}
Our contributions are summarized as follows:
\begin{itemize}[leftmargin=2em,itemsep=0.0em]
\vspace{-2mm}
\item[\ding{182}] \textbf{Temporal Threat Formulation.} We formulate unauthorized T2V
personalization as a video data protection problem and identify temporal denoising dynamics as the
key attack surface beyond static appearance learning.
\item[\ding{183}] \textbf{First Video Protection Framework.} We propose \ChronoLock,
the first protection framework for motion-centric T2V personalization, which combines diffusion
fitting, frame-relative denoising relations, and adjacent-frame variation to disrupt intra-chunk
motion adaptation.
\item[\ding{184}] \textbf{Comprehensive Evaluation.} We evaluate \ChronoLock on common T2V
backbones and two real-world video generation datasets, showing that it suppresses unauthorized
motion personalization while preserving subtle perturbations.
\end{itemize}

\vspace{-2mm}
\section{Related Work}
\vspace{-2mm}
\textbf{Text-to-Video (T2V) Generation Models.}
Recent progress in text-to-video generation has been largely driven by diffusion-based architectures
that extend image latent diffusion to spatio-temporal generation. Early systems such as Imagen
Video, Make-A-Video, and Phenaki established the feasibility of high-fidelity and long-form
text-conditioned video synthesis through cascaded diffusion, weakly supervised training, or
autoregressive modeling over video latents \citep{ho2022imagenvideo,singer2023makeavideo,villegas2023phenaki}.
Subsequent open models, including VideoCrafter1, further improved accessibility and quality for
research use \citep{chen2023videocrafter1}. Beyond generic generation, several works studied
data-efficient adaptation and personalization. Tune-A-Video showed that an image diffusion model
can be tuned into a video generator from a single clip \citep{wu2023tuneavideo}, AnimateDiff
decoupled motion modeling from personalized text-to-image backbones \citep{guo2024animatediff},
and MotionDirector explicitly customized motion concepts by updating temporal LoRA modules in a
text-to-video diffusion model \citep{zhao2024motiondirector}. These developments motivate our
threat model, in which unauthorized T2V fine-tuning is both feasible and increasingly lightweight.

\textbf{Protective Perturbations.}
A related line of work aims to protect user data from downstream model training through
imperceptible perturbations. Early methods such as Fawkes and LowKey
focused on preventing unauthorized face recognition by poisoning the training data collected from
public images \citep{shan2020fawkes,cherepanova2021lowkey}. This idea was later extended to
generative models: Glaze protected artists against style mimicry, while Nightshade introduced
prompt-specific poisoning against text-to-image generation \citep{shan2023glaze,shan2024nightshade}.
Closer to our setting, Anti-DreamBooth and Disrupting Diffusion studied how poisoned examples can
degrade diffusion-based customization or erase concept-specific attention in personalized
text-to-image models \citep{le2023antidreambooth,liu2024disrupting}. Our work differs from these
image-centric defenses in that unauthorized T2V personalization must fit not only subject
appearance but also temporal denoising dynamics, making motion consistency and cross-frame
adaptation central protection targets.

\section{Preliminaries and Problem Formulation}
\vspace{-2mm}
\subsection{Preliminaries}
\textbf{T2V Diffusion Models.} Current T2V extend latent diffusion from images to videos by denoising a
spatio-temporal latent conditioned on text. Given a training video $x$ and its text prompt $y$, a
video autoencoder first maps the video to a latent representation $z_0=\mathcal{E}(x)$, while a
text encoder produces the conditioning embedding $\tau(y)$. At diffusion step $t$, Gaussian noise
$\epsilon \sim \mathcal{N}(0, I)$ is added to the latent according to
$z_t = \sqrt{\bar{\alpha}_t} z_0 + \sqrt{1-\bar{\alpha}_t} \epsilon,~\bar{\alpha}_t = \prod_{s=1}^{t} \alpha_s$, where $\{\alpha_s\}_{s=1}^{T}$ is the noise schedule. The denoising network
$\epsilon_\theta$ is typically a 3D U-Net with down-sampling, middle, and
up-sampling blocks, where convolution layers have spatial and temporal attention to
model frame appearance and cross-frame dynamics. Its standard denoising objective is
\begin{equation}\small
\mathcal{L}_{\mathrm{den}}
\!=\!
\mathbb{E}_{(x,y), \epsilon, t}
\left[
\left\|
\epsilon \!-\! \epsilon_\theta \left( z_t, t, \tau(y) \right)
\right\|_2^2
\right]\!=\!\mathbb{E}_{(x,y), \epsilon, t}
\left[
\left\|
\epsilon \!-\! \epsilon_{\theta}\!\left(
\sqrt{\bar{\alpha}_{t}}\,\mathcal{E}(x)
\!+\! \sqrt{1-\bar{\alpha}_{t}}\,\epsilon, t, \tau(y) \right)
\right\|_2^2
\right].
\end{equation}
Compared with text-to-image generation, the T2V model must learn not only semantic and visual
content in individual frames, but also temporal coherence across the whole video.

\textbf{T2V Personalization.}
Given a small reference set $\mathcal{X}=\{(x_i, y_i)\}_{i=1}^{N}$ describing a target subject,
style, or motion pattern, T2V personalization aims to adapt a pre-trained T2V model such
that the customized concept can be generated under new text prompts while preserving the base
model's general prior. A direct approach is to fine-tune all parameters, but this is expensive for
large T2V backbones and can easily overfit when only a few videos are available. Therefore, many
methods adopt parameter-efficient adaptation and only optimize a small set of trainable parameters
while freezing the original model.
A common choice is Low-Rank Adaptation (LoRA) \citep{hu2021lora}, which parameterizes the update of a weight
matrix $W \in \mathbb{R}^{d \times k}$ as
$W' = W + \Delta W,
~\Delta W = \frac{\gamma}{r} B A$,
where $A \in \mathbb{R}^{r \times k}$ and $B \in \mathbb{R}^{d \times r}$ are trainable low-rank
matrices, $r \ll \min(d,k)$ is the adaptation rank, and $\gamma$ is a scaling factor. Denoting the
trainable parameters by $\phi$, the adaptation objective is
\begin{equation}\small
\mathcal{L}_{\mathrm{per}}(\mathcal{X};\phi)
=
\mathbb{E}_{(x,y)\sim \mathcal{X}, \epsilon, t}
\left[
\left\|
\epsilon - \epsilon_{\theta,\phi}\!\left(
\sqrt{\bar{\alpha}_{t}}\,\mathcal{E}(x)
+ \sqrt{1-\bar{\alpha}_{t}}\,\epsilon, t, \tau(y) \right)
\right\|_2^2
\right]
+ \lambda \cdot \mathcal{L}_{\mathrm{prior}},
\label{eq:training_loss}
\end{equation}
where $\theta$ is the frozen pre-trained parameters, $\mathcal{L}_{\mathrm{prior}}$
regularizes the personalized model with generic class prompts to reduce overfitting, and $\mathcal{L}_{\mathrm{per}}$ is the final training loss. In practice, spatial adapters are more effective for appearance-specific cues, whereas temporal
adapters or motion modules are better suited for capturing motion patterns. This formulation covers
common T2V personalization settings.

\vspace{-2mm}
\subsection{Problem Formulation}
\vspace{-2mm}
We consider the problem of protecting a user's videos from unauthorized text-to-video personalization.
Let $\mathcal{X}_{c}=\{(x_i,y_i)\}_{i=1}^{N}$ denote the owner's private concept set, where $x_i$ is a
clean reference video and $y_i$ is its associated text prompt. Before releasing the data, the owner
adds a small perturbation $\delta_i$ to each video and, publishes the protected, set
$\mathcal{X}_{p}=\{(\tilde{x}_i,y_i)\}_{i=1}^{N}$, where
$\tilde{x}_i = \Pi_{[0,1]}(x_i + \delta_i), ~\|\delta_i\|_{\infty} \leq \varepsilon$, and $\Pi_{[0,1]}(\cdot)$ projects pixel values back to the valid range. For simplicity, unprotected
samples can be treated as the special case $\delta_i=0$.

An unauthorized model trainer then collects $\mathcal{X}_{p}$ and uses it to personalize a
pre-trained T2V diffusion model. Given the frozen backbone parameters $\theta$ and the
trainable adaptation parameters $\phi$ (\textit{e.g.,} LoRA), the personalization objective of
the trainer can be formulated as ($\mathcal{L}_{\mathrm{per}}$ defined in Eq.~\eqref{eq:training_loss}):
\begin{equation}\small
\phi^{\star}(\Delta)
\in
\arg\min_{\phi}
\mathcal{L}_{\mathrm{per}}(\mathcal{X}_{p};\phi),\quad\text{where}\quad\Delta=\{\delta_i\}_{i=1}^{N}.
\end{equation}
The defender aims to craft a perturbation set $\Delta$ such that the model personalized on
$\mathcal{X}_{p}$ becomes unusable for downstream generation, while the released videos remain
visually close to the originals. We formulate the protection objective as the following bi-level
optimization:
\vspace{-1mm}
\begin{equation}\small
\Delta^{\star}
\in
\arg\max_{\Delta}
\mathcal{L}_{\mathrm{p}}\!\left(\epsilon_{\theta,\phi^{\star}(\Delta)};\mathcal{X}_{c}\right)
\quad
\text{s.t.}
\quad
\forall~i:~\|\delta_i\|_{\infty} \leq \varepsilon.
\end{equation}
The outer objective $\mathcal{L}_{\mathrm{p}}$ denotes the protection objective and measures the
failure of the personalized T2V model. Because there is no single scalar metric that fully captures
personalization quality.

\begin{figure*}
\centering
\includegraphics[width=\textwidth]{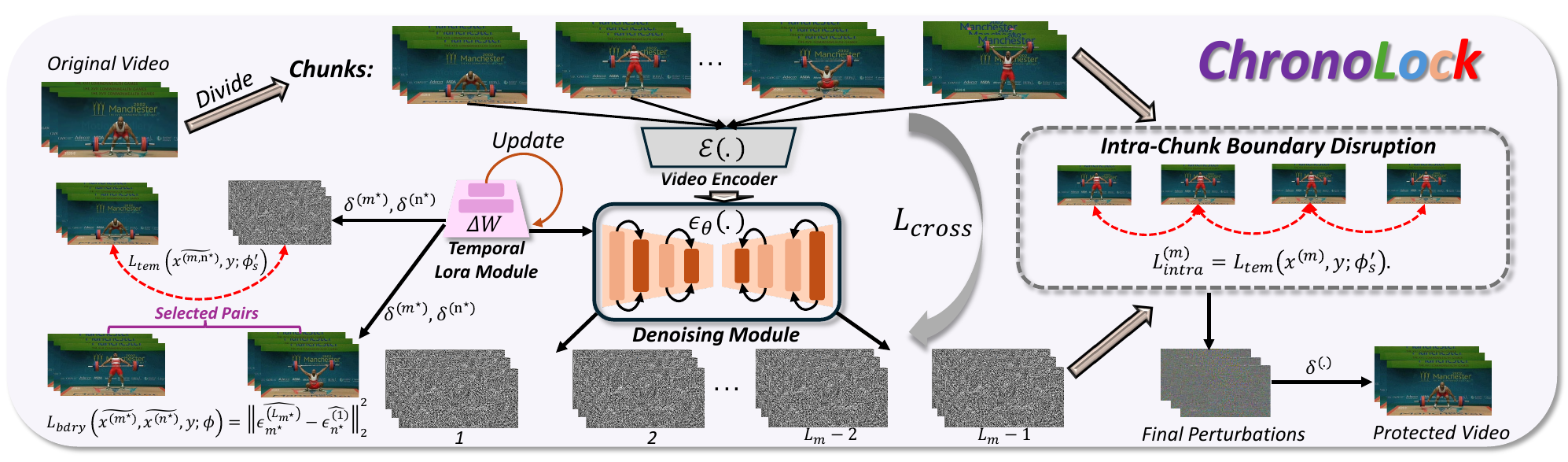} 
\vspace{-7mm}
\caption{Overview of \ChronoLock. The defender releases visually similar protected videos whose
temporal denoising evidence becomes unreliable for unauthorized T2V personalization. \ChronoLock
first corrupts local intra-chunk motion fitting and then enlarges boundary-level continuity mismatch
to degrade personalized motion personalization.}
\label{Figure:overview}
\vspace{-3mm}
\end{figure*}

\vspace{-2mm}
\section{\ChronoLock: Temporal-aware Protection Framework}
\vspace{-2mm}
\textbf{Overview.}
Fig.~\ref{Figure:overview} illustrates the pipeline of \ChronoLock. Specifically, given a clean
video-prompt pair, \ChronoLock optimizes imperceptible perturbations against the temporal
fine-tuning dynamics of a T2V diffusion model. We first instantiate motion-centric personalization
with temporal LoRA adapters \citep{hu2021lora,zhao2024motiondirector} and define the temporal
diffusion objective used for protection ($\triangleright$ Section~\ref{sec:temporal-objective}).
\ChronoLock then performs Intra-Chunk Temporal Disruption to corrupt local motion evidence
($\triangleright$ Section~\ref{sec:intra-disruption}), followed by Inter-Chunk Boundary
Disruption to enlarge the most vulnerable temporal transition
($\triangleright$ Section~\ref{sec:boundary-disruption}).
Finally, we combine these objectives into the overall optimization with Algorithm~\ref{alg:chronolock}
($\triangleright$ Section~\ref{sec:overall-optimization}). The final protected video remains
visually close to the original but supplies unreliable temporal evidence for unauthorized
personalization. 

\vspace{-1mm}
\subsection{Temporal Diffusion Objective}
\label{sec:temporal-objective}
\vspace{-1mm}
\textbf{Temporal Adaptation Target.}
Video personalization methods often adapt temporal or motion-specific modules to capture dynamics
while preserving the spatial prior of the base generator \citep{guo2024animatediff,zhao2024motiondirector}.
Following this practice, we freeze the VAE, text encoder, and backbone U-Net, and update only the
temporal LoRA parameters $\phi$. Given a video $x \in \mathbb{R}^{F \times 3 \times H \times W}$
with prompt $y$, where $F$, $H$, and $W$ denote the number of frames, height, and width, we
partition it into $M$ contiguous chunks,
\vspace{-1mm}
\begin{equation}\small
x = [x^{(1)}, x^{(2)}, \ldots, x^{(M)}],
\qquad
x^{(m)} \in \mathbb{R}^{L_m \times 3 \times H \times W},
\qquad
\sum_{m=1}^{M} L_m = F.
\end{equation}
Here $L_m$ denotes the number of frames in chunk $m$. We optimize chunk-wise perturbations
$\Delta = \{\delta^{(m)}\}_{m=1}^{M}$, where $\delta^{(m)}$ denotes the perturbation added to
chunk $x^{(m)}$, under the same
$\ell_{\infty}$ budget as in the problem formulation, which enables local temporal losses while
preserving the original ordering for cross-chunk boundary optimization.

\textbf{Temporal Denoising Setup.}
Given the video encoder $\mathcal{E}$, for chunk $x^{(m)}$, let $z_0^{(m)}=\mathcal{E}(x^{(m)})$ be its latent representation and $z_t^{(m)} = \sqrt{\bar{\alpha}_t} z_0^{(m)} + \sqrt{1-\bar{\alpha}_t}\epsilon$ be the corresponding noisy latent at timestep $t$, where $\epsilon$ denotes sampled Gaussian noise. We denote the predicted noise by
\vspace{-1mm}
\begin{equation}\small
\hat{\epsilon}^{(m)} = \epsilon_{\theta,\phi}\!\left(z_t^{(m)}, t, \tau(y)\right).
\end{equation}
For simplicity, let $\hat{\epsilon}^{(m,\ell)}$ and $\epsilon^{(m,\ell)}$ be the
predicted and sampled noise slices at frame index $\ell \in \{1,\dots,L_m\}$.
We keep the protection objective tied to diffusion training with the fitting term:
\vspace{-1mm}
\begin{equation}\small
\mathcal{L}_{\mathrm{fit}}(x^{(m)}, y; \phi)
=
\mathbb{E}_{\epsilon,t}
\left[
\left\|
\epsilon - \epsilon_{\theta,\phi}\!\left(z_t^{(m)}, t, \tau(y)\right)
\right\|_2^2
\right],
\end{equation}
which provides the denoising signal that an adapter would normally
minimize during personalization.

\textbf{Frame-Relative Relation.}
The fitting term can be dominated by shared per-frame appearance. To expose the motion signal, we
compare each frame to a common reference slice. Sampling
$r \sim \mathrm{Unif}(\{1,\ldots,L_m\})$ and using $\beta\ge0$ to denote the reference-slice
weight, we then define
\begin{equation}\small
\!\!\mathcal{L}_{\mathrm{rel}}(x^{(m)}, y; \phi)
\!=\!
\mathbb{E}_{\epsilon,t,r}\!\!
\left[\!
\frac{1}{L_m}
\sum_{\ell=1}^{L_m}
\left\|
\alpha \hat{\epsilon}^{(m,\ell)} \!-\! \beta \hat{\epsilon}^{(m,r)}
\!-\! \left(\alpha \epsilon^{(m,\ell)} \!-\! \beta \epsilon^{(m,r)}\right)
\right\|_2^2
\!\right],~\alpha\!=\!\sqrt{1\!+\!\beta^2}.
\end{equation}
The shared reference cancels chunk-level static
components, while the scaling keeps the linear combination numerically controlled. Minimizing
$\mathcal{L}_{\mathrm{rel}}$ encourages stable frame-relative denoising, while maximizing it makes such
relations difficult to recover from protected videos.

\textbf{Adjacent-Frame Variation.}
We further expose local temporal inconsistency by measuring how predicted denoising slices change
between neighboring frames. When $L_m = 1$, we set $\mathcal{L}_{\mathrm{var}}(x^{(m)}, y; \phi)=0$. For $L_m\ge2$, we define
\begin{equation}\small
\mathcal{L}_{\mathrm{var}}(x^{(m)}, y; \phi)
=
\mathbb{E}_{\epsilon,t}
\left[
\frac{1}{L_m-1}
\sum_{\ell=1}^{L_m-1}
\left\|
\hat{\epsilon}^{(m,\ell+1)} - \hat{\epsilon}^{(m,\ell)}
\right\|_2^2
\right].
\end{equation}
\begin{property}[Temporal Variation Characterization]
For any fixed chunk $x^{(m)}$ with $L_m \ge 2$ and prompt $y$,
\begin{equation}\small
\mathcal{L}_{\mathrm{var}}(x^{(m)}, y; \phi)=0
\iff
\hat{\epsilon}^{(m,\ell+1)}=\hat{\epsilon}^{(m,\ell)},
\quad
\forall \ell \in \{1,\dots,L_m-1\},
\end{equation}
Thus, $\mathcal{L}_{\mathrm{var}}$ gets lower when the
predicted denoising trajectory is temporally constant.
\end{property}

\textbf{Temporal Diffusion Objective.} The final temporal objective used for protection is
\begin{equation}\small
\mathcal{L}_{\mathrm{tem}}(x^{(m)}, y; \phi)
=
\mathcal{L}_{\mathrm{fit}}(x^{(m)}, y; \phi)
+ \mathcal{L}_{\mathrm{rel}}(x^{(m)}, y; \phi)
+ \lambda_c \cdot \mathcal{L}_{\mathrm{var}}(x^{(m)}, y; \phi),
\end{equation}
where $\lambda_c>0$ controls the adjacent-frame term. Minimizing $\mathcal{L}_{\mathrm{tem}}$ models
temporal personalization, while maximizing it attacks the same motion-learning signal.

\subsection{Intra-Chunk Temporal Disruption}
\label{sec:intra-disruption}

\textbf{Intra-Chunk Objective.}
The temporal objective above exposes the short-range motion evidence that a temporal adapter would
normally fit. \ChronoLock therefore makes each local chunk difficult for temporal LoRA adaptation by
maximizing the objective that the adapter would otherwise minimize. For each protected chunk
$\tilde{x}^{(m)}=x^{(m)}+\delta^{(m)}$, where $\phi_s'$ denotes the temporary adapter obtained at
outer step $s$, the intra-chunk objective is
\begin{equation}\small
\mathcal{L}_{\mathrm{intra}}^{(m)}
=
\mathcal{L}_{\mathrm{tem}}\!\left(\tilde{x}^{(m)}, y; \phi_s'\right).
\label{eq:intra}
\end{equation}

\subsection{Inter-Chunk Boundary Disruption}
\label{sec:boundary-disruption}

\textbf{Adaptive Boundary Selection.}
Local disruption weakens short-range motion evidence, but a video generator can still compose
locally plausible segments if their boundary predictions remain compatible. Inter-Chunk Boundary
Disruption therefore targets the transition that currently carries the largest denoising
mismatch. Let $\tilde{x}^{(m)}=x^{(m)}+\delta^{(m)}$ be the current protected chunk, and denote the
first and last denoising slices by $\hat{\epsilon}_{m}^{(1)}$ and
$\hat{\epsilon}_{m}^{(L_m)}$, respectively. These slices are computed from $\tilde{x}^{(m)}$ using
the current adapter, timestep, and sampled diffusion noise. For $M\ge2$, we score each pair by
\begin{equation}\small
S(m,n)
=
\left\|
\hat{\epsilon}_{m}^{(L_m)} - \hat{\epsilon}_{n}^{(1)}
\right\|_2^2,
\qquad
1 \le m < n \le M,
\end{equation}
where $S(m,n)$ denotes the boundary mismatch score between chunks $m$ and $n$. We choose
\begin{equation}\small
(m^{\star}, n^{\star})
\in
\arg\max_{1 \le m < n \le M} S(m,n).
\end{equation}

\textbf{Pairwise Boundary Objective.}
After selecting $(m^\star,n^\star)$, we again obtain a temporary adapter $\phi_s'$ through $K$ clean
inner steps on the selected pair. The boundary term is
\begin{equation}\small
\mathcal{L}_{\mathrm{bdry}}
\left(
\tilde{x}^{(m^{\star})},
\tilde{x}^{(n^{\star})},
y;\phi
\right)
=
\left\|
\hat{\epsilon}_{m^{\star}}^{(L_{m^{\star}})}
- \hat{\epsilon}_{n^{\star}}^{(1)}
\right\|_2^2.
\end{equation}
The selected-pair objective is
\begin{equation}\small
\mathcal{L}_{\mathrm{cross}}
=
\mathcal{L}_{\mathrm{tem}}\!\left(\tilde{x}^{(m^{\star})}, y; \phi_s'\right)
+ \mathcal{L}_{\mathrm{tem}}\!\left(\tilde{x}^{(n^{\star})}, y; \phi_s'\right)
+ \lambda_b
\mathcal{L}_{\mathrm{bdry}}
\left(
\tilde{x}^{(m^{\star})},
\tilde{x}^{(n^{\star})},
y;\phi_s'
\right),
\label{eq:inter}
\end{equation}
where $\lambda_b>0$ controls the boundary signal.

\begin{proposition}[Adaptive Boundary Focus]
For fixed current chunks and temporary parameters $\phi_s'$, the selected pair satisfies
\begin{equation}\small
(m^{\star}, n^{\star})
\in
\arg\max_{1 \le m < n \le M}
\mathcal{L}_{\mathrm{bdry}}
\left(
\tilde{x}^{(m)},
\tilde{x}^{(n)},
y;\phi_s'
\right).
\end{equation}
\ChronoLock allocates each global update to the boundary with the largest
denoising mismatch.    
\end{proposition}

\subsection{Overall Optimization}
\label{sec:overall-optimization}

\begin{figure}[t]
\begin{algorithm}[H]\small
    \caption{Workflow of \ChronoLock}
    \label{alg:chronolock}
    \LinesNotNumbered
    \SetKwInOut{Input}{Input}
    \SetKwInOut{Output}{Output}
    \DontPrintSemicolon

    \Input{Clean video $x$, prompt $y$, budget $\varepsilon$, chunk size, temporal LoRA parameters $\phi$,
    transformation distribution $\mathcal{T}_{\mathrm{aug}}$, intra-chunk steps $S$, cross-chunk steps $S_g$,
    inner steps $K$, synchronization interval $K_{\mathrm{sync}}$, weights $\lambda_c,\lambda_b$.}
    \Output{Protected video $\tilde{x}$}

    Split $x$ into chunks $\{x^{(m)}\}_{m=1}^{M}$ and initialize $\{\delta^{(m)}\}_{m=1}^{M}$\;

    \mycomment{Phase 1: intra-chunk temporal disruption}\\
    \For{$m=1,\ldots,M$}{
        \For{$s=1,\ldots,S$}{
            $\tilde{x}^{(m)}\leftarrow x^{(m)}+\delta^{(m)}$; snapshot $\phi_s$\;
            Obtain $\phi_s'$ by taking $K$ inner steps on $\tilde{x}^{(m)}$ to minimize the temporal personalization loss\;
            Sample $g\sim\mathcal{T}_{\mathrm{aug}}$\;
            $\delta^{(m)} \leftarrow
            \Pi_{\mathcal{C}(x^{(m)})}\!\left[
            \delta^{(m)}+\rho\,\mathrm{sign}\!\left(
            \nabla_{\delta^{(m)}}\mathcal{L}_{\mathrm{intra}}^{(m)}
            (g(\tilde{x}^{(m)}),y;\phi_s')
            \right)\right]$\;
            Restore $\phi_s$ and synchronize the running LoRA state every $K_{\mathrm{sync}}$ steps\;
        }
    }

    \mycomment{Phase 2: adaptive cross-chunk boundary disruption}\\
    \For{$s=1,\ldots,S_g$}{
        Select $(m^\star,n^\star)\leftarrow\arg\max_{m<n}S(m,n)$ over current protected chunks\;
        Snapshot $\phi_s$ and obtain $\phi_s'$ by taking $K$ inner steps on the selected protected pair\;
        Sample $g_m,g_n\sim\mathcal{T}_{\mathrm{aug}}$\;
        Update only $\delta^{(m^\star)}$ and $\delta^{(n^\star)}$ by projected gradient ascent on
        $\mathcal{L}_{\mathrm{cross}}(g_m(\tilde{x}^{(m^\star)}),g_n(\tilde{x}^{(n^\star)}),y;\phi_s')$\;
        Restore $\phi_s$ and keep all unselected chunks fixed\;
    }

    \Return $\tilde{x}\leftarrow[x^{(1)}+\delta^{(1)},\ldots,x^{(M)}+\delta^{(M)}]$\;
\end{algorithm}
\vspace{-4mm}
\end{figure}

\textbf{Two-Stage Perturbation Optimization.}
\ChronoLock implements the protection process as a two-stage projected optimization rather than
optimizing a single joint objective over the whole video. In the first stage, it updates all chunks
independently. For each chunk, the algorithm first simulates the unauthorized personalization
process on the current protected chunk $\tilde{x}^{(m)}=x^{(m)}+\delta^{(m)}$: starting from the
current temporal LoRA parameters $\phi_s$, it takes $K$ inner optimization steps to minimize the
temporal personalization loss, which is the chunk-wise instantiation of Eq.~\ref{eq:training_loss}.
This produces a temporary adapter $\phi_s'$ that approximates how an unauthorized trainer would
adapt to the protected video. \ChronoLock then performs PGD~\citep{madry2018towards} on $\delta^{(m)}$ to maximize the intra-chunk objective
$\mathcal{L}_{\mathrm{intra}}^{(m)}$ in Eq.~\ref{eq:intra}. This stage corrupts local temporal
denoising evidence in every short video segment.

In the second stage, \ChronoLock further refines the protected video by targeting cross-chunk
motion continuity. At each global PGD step, it evaluates the current boundary mismatch score
$S(m,n)$ over all chunk pairs and selects the pair $(m^\star,n^\star)$ with the largest mismatch.
The temporary adapter $\phi_s'$ is again obtained by minimizing the temporal personalization loss
on the selected protected pair, and only the perturbations of this pair are updated by projected
gradient ascent on $\mathcal{L}_{\mathrm{cross}}$ in Eq.~\ref{eq:inter}. All other chunk
perturbations remain fixed. This adaptive update concentrates the optimization on the currently
most vulnerable temporal transition and enlarges the boundary-level denoising mismatch.

\textbf{Projection and Transformation Sampling.}
For a clean chunk $x^{(m)}$, each perturbation update is projected onto the feasible set
\vspace{-1mm}
\begin{equation}\small
\mathcal{C}(x^{(m)})
=
\left\{
\delta :
\|\delta\|_{\infty} \le \varepsilon,\ 
x^{(m)} + \delta \in [0,1]^{L_m \times 3 \times H \times W}
\right\}.
\end{equation}
\vspace{-1mm}
This projection keeps the protected chunk within the valid video range and the prescribed
perturbation budget. To improve robustness against simple preprocessing, the outer perturbation
update is evaluated on transformed views sampled from $\mathcal{T}_{\mathrm{aug}}$, such as Gaussian
blur and horizontal flip. The temporary adapter $\phi_s'$ is discarded after each outer update, and
the running LoRA state is periodically synchronized with the current protected video segment every
$K_{\mathrm{sync}}$ steps.

\begin{discussionbox}
\small
\textbf{Discussion.}
\textit{The two-stage optimization couples two temporal failure modes for T2V personalization learning.
Intra-chunk maximization makes every local motion segment hard to fit, while adaptive cross-chunk
maximization enlarges the most inconsistent transition across chunks. Thus, projected ascent attacks
both short-range denoising evidence and long-range motion continuity under the same perturbation
budget.}
\end{discussionbox}
\vspace{-2mm}
\section{Experiments}
\vspace{-2mm}
\textbf{18.}
The \ChronoLock protection is implemented against MotionDirector temporal personalization~\citep{zhao2024motiondirector}. We initialize Temporal LoRA on \texttt{TransformerTemporalModel} with rank 16. All videos are resized to $576 \times 320$ and split into 5-frame chunks. Following Algorithm~\ref{alg:chronolock}, we set $\varepsilon=0.05$, intra-chunk PGD steps $S=20$, cross-chunk PGD steps $S_g=20$, inner steps $K=5$, inner learning rate $1\mathrm{e}{-4}$, synchronization interval $K_{\mathrm{sync}}=10$, surrogate learning rate $1\mathrm{e}{-5}$, and PGD step size $\rho=2.5\varepsilon/S$. The loss weights are set to $\lambda_c=0.2$ and $\lambda_b=0.5$. 
Phase 1 updates each chunk by maximizing $\mathcal{L}_{\mathrm{intra}}^{(m)}$, while Phase 2 updates only the selected chunk pair with the largest boundary score by maximizing $\mathcal{L}_{\mathrm{cross}}$.

\textbf{Datasets \& Models.}
We use 2 real-world human action datasets, UCF Sports~\citep{rodriguez2008action} and HMDB51~\citep{kuehne2011hmdb}.
For models, we choose 2 widely used T2V diffusion backbones, ZeroScope~\citep{sterling2023zeroscope} and ModelScope~\citep{wang2023modelscope}, and use MotionDirector~\citep{zhao2024motiondirector} with Temporal LoRA for the downstream motion customization.

\textbf{Metrics.}
For each trained MotionDirector model and test prompt, we
generate multiple videos and evaluate whether the customized model preserves the protected motion.
We report two benchmark-sourced automatic metrics: motion smoothness (MS)
from VBench~\citep{huang2024vbench} and temporal
consistency (TC) from Video-Bench~\citep{han2025video}.
MS and TC measure temporal smoothness and frame-level consistency, respectively.
Lower MS and TC values indicate weaker temporal quality and stronger protection.
We further conduct human evaluation along three axes: text alignment (TA), temporal
consistency (TCons), and motion fidelity (MF). Evaluation details are given in the Appendix~\ref{HumanEvaluationDetails}.

\vspace{-2mm}
\subsection{Main Results}
\vspace{-2mm}

\begin{table}[H]
\centering
\setlength{\tabcolsep}{2.pt}
\renewcommand{\arraystretch}{1.0}
\caption{Results of different schemes under the unified evaluation setting. We report automatic metrics MS and TC, as well as human preference scores on TA, TCons, and MF. For each human-evaluation metric, we additionally report the P-O/C-O ratio, computed as $(\ChronoLock-\mathrm{Original})/(\mathrm{Clean}-\mathrm{Original})$, where P denotes \ChronoLock. The best performances are highlighted in bold.}
\label{tab:main_results_n}

\resizebox{\textwidth}{!}{%
\begin{tabular}{c|c|c||cc|cc|cc|cc}
\hline
\multirow{2}{*}{\textbf{Dataset}}
& \multirow{2}{*}{\textbf{Model}}
& \multirow{2}{*}{\textbf{Scheme}}
& \multicolumn{2}{c|}{\cellcolor{gray!20}\textbf{Automatic Evaluation}}
& \multicolumn{6}{c}{\cellcolor{gray!20}\textbf{Human Evaluation}} \\

&
&
& \cellcolor{gray!20}\textbf{MS}$\downarrow$
& \cellcolor{gray!20}\textbf{TC}$\downarrow$
& \cellcolor{gray!20}\textbf{TA}$\downarrow$
& \cellcolor{gray!20}\textbf{P-O/C-O}$\downarrow$
& \cellcolor{gray!20}\textbf{TCons}$\downarrow$
& \cellcolor{gray!20}\textbf{P-O/C-O}$\downarrow$
& \cellcolor{gray!20}\textbf{MF}$\downarrow$
& \cellcolor{gray!20}\textbf{P-O/C-O}$\downarrow$ \\
\hline\hline

\multirow[c]{6}{*}{\rotatebox{90}{\textbf{UCF}}}
& \multirow[c]{3}{*}{ZeroScope}
& Original
& 0.989
& 4.324
& 38.720
&
& 45.860
&
& 27.730
& \\

& &
\cellcolor{gray!10} Clean
& \cellcolor{gray!10} 0.995
& \cellcolor{gray!10} 4.662
& \cellcolor{gray!10} 70.000
& \cellcolor{gray!10}
& \cellcolor{gray!10} 69.090
& \cellcolor{gray!10}
& \cellcolor{gray!10} 68.360
& \cellcolor{gray!10} \\

& &
\ChronoLock
& \textbf{0.980}
& \textbf{4.149}
& \textbf{30.000}
& \textbf{-0.279}
& \textbf{30.910}
& \textbf{-0.644}
& \textbf{31.640}
& \textbf{0.096} \\

\cline{2-11}

& \multirow[c]{3}{*}{ModelScope}
& \cellcolor{gray!10} Original
& \cellcolor{gray!10} 0.977
& \cellcolor{gray!10} 4.176
& \cellcolor{gray!10} 35.960
& \cellcolor{gray!10}
& \cellcolor{gray!10} 42.980
& \cellcolor{gray!10}
& \cellcolor{gray!10} 29.550
& \cellcolor{gray!10} \\

& &
Clean
& 0.994
& 4.540
& 69.330
&
& 70.610
&
& 68.860
& \\

& &
\cellcolor{gray!10}\ChronoLock
& \cellcolor{gray!10}\textbf{0.985}
& \cellcolor{gray!10}\textbf{4.294}
& \cellcolor{gray!10}\textbf{30.670}
& \cellcolor{gray!10}\textbf{-0.159}
& \cellcolor{gray!10}\textbf{29.390}
& \cellcolor{gray!10}\textbf{-0.492}
& \cellcolor{gray!10}\textbf{31.140}
& \cellcolor{gray!10}\textbf{0.040} \\

\hline

\multirow[c]{6}{*}{\rotatebox{90}{\textbf{HMDB51}}}
& \multirow[c]{3}{*}{ZeroScope}
& Original
& 0.991
& 4.300
& 47.990
&
& 41.280
&
& 24.860
& \\

& &
\cellcolor{gray!10} Clean
& \cellcolor{gray!10} 0.995
& \cellcolor{gray!10} 4.538
& \cellcolor{gray!10} 66.940
& \cellcolor{gray!10}
& \cellcolor{gray!10} 67.990
& \cellcolor{gray!10}
& \cellcolor{gray!10} 70.270
& \cellcolor{gray!10} \\

& &
\ChronoLock
& \textbf{0.988}
& \textbf{4.333}
& \textbf{33.060}
& \textbf{-0.788}
& \textbf{32.010}
& \textbf{-0.347}
& \textbf{29.730}
& \textbf{0.107} \\

\cline{2-11}

& \multirow[c]{3}{*}{ModelScope}
& \cellcolor{gray!10} Original
& \cellcolor{gray!10} 0.983
& \cellcolor{gray!10} 4.139
& \cellcolor{gray!10} 45.900
& \cellcolor{gray!10}
& \cellcolor{gray!10} 41.250
& \cellcolor{gray!10}
& \cellcolor{gray!10} 30.750
& \cellcolor{gray!10} \\

& &
Clean
& 0.993
& 4.765
& 68.670
&
& 59.070
&
& 62.950
& \\

& &
\cellcolor{gray!10}\ChronoLock
& \cellcolor{gray!10}\textbf{0.988}
& \cellcolor{gray!10}\textbf{4.389}
& \cellcolor{gray!10}\textbf{31.330}
& \cellcolor{gray!10}\textbf{-0.640}
& \cellcolor{gray!10}\textbf{40.930}
& \cellcolor{gray!10}\textbf{-0.018}
& \cellcolor{gray!10}\textbf{37.050}
& \cellcolor{gray!10}\textbf{0.196} \\

\hline
\end{tabular}
}
\vspace{-4mm}
\end{table}

\begin{figure*}
\centering
\includegraphics[width=\textwidth]{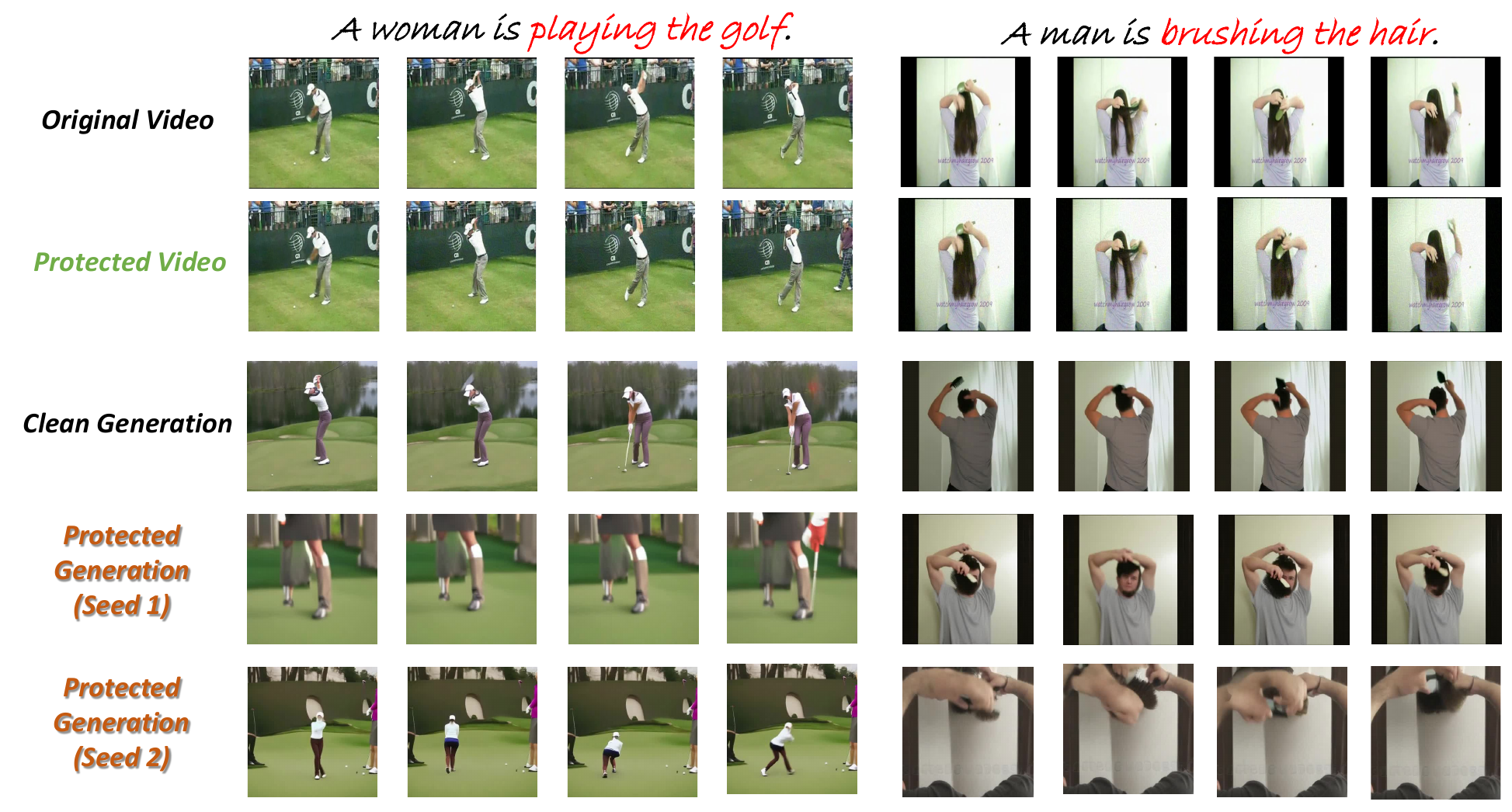} 
\vspace{-6mm}
\caption{Qualitative defense results for two users' videos.}
\label{Figure:vis}
\vspace{-2mm}
\end{figure*}

\textbf{Results under the Unified Evaluation.}
As shown in Table~\ref{tab:main_results_n}, \ChronoLock
consistently pushes unauthorized motion personalization toward low-utility
generations. Averaged over the human-evaluation axes and all dataset-backbone
settings, the Clean baseline achieves 67.68\% preference, while \ChronoLock
suppresses this to 32.32\%. The automatic evaluation supports the same
conclusion from complementary temporal perspectives. Lower MS scores indicate
that \ChronoLock degrades motion smoothness, while lower TC scores show that the
generated videos lose temporal consistency across frames.
Our findings indicate that \ChronoLock exploits temporal motion-learning
vulnerabilities rather than dataset-specific artifacts. The consistent results
across UCF and HMDB51 show that \ChronoLock provides a broadly applicable
protection strategy that is not sensitive to the action category set or the
underlying T2V generation backbone. Some visual results are in Fig.~\ref{Figure:vis}.

\begin{table}[H]
\centering
\footnotesize
\setlength{\tabcolsep}{4.5pt}
\renewcommand{\arraystretch}{1.08}
\caption{Resistance to defenses.}
\label{tab:defense_robustness}
\resizebox{0.86\textwidth}{!}{%
\begin{tabular}{c|c||cc|ccc}
\hline
\rowcolor{gray!20}
\textbf{Dataset}
& \textbf{Defense}
& \multicolumn{2}{c|}{\textbf{Automatic Metrics}}
& \multicolumn{3}{c}{\textbf{Human Preference (\%)}} \\

\rowcolor{gray!20}
\multicolumn{1}{l|}{$\downarrow$}
& \multicolumn{1}{l||}{$\downarrow$}
& \textbf{MS}$\downarrow$
& \textbf{TC}$\downarrow$
& \textbf{TA}$\downarrow$
& \textbf{TCons}$\downarrow$
& \textbf{MF}$\downarrow$ \\
\hline\hline

\multirow[c]{4}{*}{\rotatebox{90}{\textbf{UCF}}}
& No Defense
& 0.980
& 4.149
& 30.000
& 30.910
& 31.640 \\
\cdashline{2-7}

& \cellcolor{gray!10} Gaussian Blur
& \cellcolor{gray!10} 0.989
& \cellcolor{gray!10} 4.475
& \cellcolor{gray!10} 35.797
& \cellcolor{gray!10} 41.481
& \cellcolor{gray!10} 42.712 \\

& Horizontal Flip
& 0.983
& 4.225
& 30.390
& 31.870
& 33.060 \\

& \cellcolor{gray!10} Blur + HFlip
& \cellcolor{gray!10} 0.992
& \cellcolor{gray!10} 4.500
& \cellcolor{gray!10} 37.800
& \cellcolor{gray!10} 44.160
& \cellcolor{gray!10} 45.000 \\
\hline

\multirow[c]{4}{*}{\rotatebox{90}{\textbf{HMDB51}}}
& No Defense
& 0.988
& 4.333
& 33.060
& 32.010
& 29.730 \\
\cdashline{2-7}

& \cellcolor{gray!10} Gaussian Blur
& \cellcolor{gray!10} 0.992
& \cellcolor{gray!10} 4.407
& \cellcolor{gray!10} 35.950
& \cellcolor{gray!10} 36.950
& \cellcolor{gray!10} 37.920 \\

& Horizontal Flip
& 0.988
& 4.369
& 34.640
& 33.020
& 29.900 \\

& \cellcolor{gray!10} Blur + HFlip
& \cellcolor{gray!10} 0.994
& \cellcolor{gray!10} 4.463
& \cellcolor{gray!10} 37.710
& \cellcolor{gray!10} 38.400
& \cellcolor{gray!10} 39.570 \\
\hline
\end{tabular}
}
\end{table}
\textbf{Resistance to defenses.}
To evaluate the resistance of our proposed protection, we employ common video
augmentation and purification operations as defense mechanisms, attempting to
neutralize the injected perturbations and restore unauthorized motion
personalization. We test three transformation defenses, including Gaussian blur,
horizontal flip, and their composition. Results in Table~\ref{tab:defense_robustness} reveal that while applying
these transformations can yield a marginal recovery in automatic and 
human
evaluation, the overall improvement remains limited and fails to recover the
semantic and motion utility of the protected videos. Among the evaluated
defenses, Blur+HFlip emerges as the relatively most effective strategy,
consistently producing stronger recovery than single transformations.
Nevertheless, the recovered utility remains far below the clean personalization
level, showing that \ChronoLock preserves effective protection under diverse pre-processing methods.

\begin{table}[H]
\centering
\scriptsize
\setlength{\tabcolsep}{3.2pt}
\renewcommand{\arraystretch}{1.08}
\caption{Mismatch evaluation across datasets: cross-model transfer and prompt mismatch results.}

\label{tab:mismatch_results}

\resizebox{\textwidth}{!}{%
\begin{tabular}{
>{\centering\arraybackslash}p{1.65cm}|
>{\centering\arraybackslash}p{1.35cm}|
>{\centering\arraybackslash}p{2.05cm}|
>{\centering\arraybackslash}p{2.05cm}||
*{2}{>{\centering\arraybackslash}p{1.25cm}}|
*{3}{>{\centering\arraybackslash}p{1.25cm}}
}
\hline
\rowcolor{gray!20}
\textbf{Mismatch Type}
& \textbf{Dataset}
& \textbf{Protect Model}
& \textbf{Train Model}
& \multicolumn{2}{c|}{\textbf{Automatic Evaluation}}
& \multicolumn{3}{c}{\textbf{Human Evaluation}} \\

\rowcolor{gray!20}
\multicolumn{1}{l|}{$\downarrow$}
& \multicolumn{1}{l|}{$\downarrow$}
& \multicolumn{1}{l|}{$\downarrow$}
& \multicolumn{1}{l||}{$\downarrow$}
& \textbf{MS}$\downarrow$
& \textbf{TC}$\downarrow$
& \textbf{TA}$\downarrow$
& \textbf{TCons}$\downarrow$
& \textbf{MF}$\downarrow$ \\
\hline\hline

\multirow[c]{4}{*}{\makecell[c]{\textbf{Model}\\\textbf{mismatch}}}
& \multirow[c]{2}{*}{\textbf{UCF}}
& ZeroScope
& ModelScope
& 0.990
& 4.176
& 32.650
& 32.570
& \textbf{33.320} \\

& 
& ModelScope
& ZeroScope
& \textbf{0.967}
& \textbf{4.137}
& \textbf{29.880}
& \textbf{28.920}
& 36.730 \\

\cline{2-9}

& \multirow[c]{2}{*}{\textbf{HMDB51}}
& ZeroScope
& ModelScope
& \textbf{0.991}
& \textbf{4.319}
& 35.750
& \textbf{31.280}
& 36.500 \\

&
& ModelScope
& ZeroScope
& 0.991
& 4.514
& \textbf{33.640}
& 36.660
& \textbf{33.030} \\

\hline\hline

\multirow[c]{2}{*}{\makecell[c]{\textbf{Prompt}\\\textbf{mismatch}}}
& \textbf{UCF}
& \multicolumn{2}{c||}{Prompt mismatch}
& \textbf{0.981}
& 4.122
& \textbf{34.680}
& 36.225
& 35.148 \\

& \cellcolor{gray!10}\textbf{HMDB51}
& \multicolumn{2}{c||}{\cellcolor{gray!10} Prompt mismatch}
& \cellcolor{gray!10} 0.986
& \cellcolor{gray!10} \textbf{4.000}
& \cellcolor{gray!10} 35.157
& \cellcolor{gray!10} \textbf{33.163}
& \cellcolor{gray!10} \textbf{33.210} \\

\hline
\end{tabular}
}
\vspace{-4mm}
\end{table}

\begin{table}[H]
\centering
\footnotesize
\setlength{\tabcolsep}{4.0pt}
\renewcommand{\arraystretch}{1.08}
\caption{Impact of perturbation budget $\epsilon$: automatic metrics and human preference scores.}
\label{tab:epsilon_sensitivity}

\resizebox{\textwidth}{!}{%
\begin{tabular}{c|l||ccccc|ccccc}
\hline
\rowcolor{gray!20}
\multicolumn{2}{l||}{\textbf{Dataset} $\rightarrow$}
& \multicolumn{5}{c|}{\textbf{UCF}}
& \multicolumn{5}{c}{\textbf{HMDB51}} \\

\rowcolor{gray!20}
\multicolumn{2}{l||}{\textbf{Evaluation} $\downarrow$, \textbf{Budget} $\rightarrow$}
& $\epsilon=0.01$
& $\epsilon=0.02$
& $\epsilon=0.03$
& $\epsilon=0.04$
& $\epsilon=0.05$
& $\epsilon=0.01$
& $\epsilon=0.02$
& $\epsilon=0.03$
& $\epsilon=0.04$
& $\epsilon=0.05$ \\
\hline\hline

\multirow[c]{2}{*}{\makecell[c]{Automatic\\Evaluation}}
& MS$\downarrow$
& 0.992
& 0.990
& 0.987
& \textbf{0.979}
& 0.980
& 0.995
& 0.994 
& 0.991 
& 0.989
& \textbf{0.988} \\

& \cellcolor{gray!10} TC$\downarrow$
& \cellcolor{gray!10} 4.257
& \cellcolor{gray!10} \textbf{4.132}
& \cellcolor{gray!10} 4.328
& \cellcolor{gray!10} 4.164
& \cellcolor{gray!10} 4.149
& \cellcolor{gray!10} 4.457
& \cellcolor{gray!10} 4.280
& \cellcolor{gray!10} 4.302
& \cellcolor{gray!10} \textbf{4.188}
& \cellcolor{gray!10} 4.333 \\

\hline

\multirow[c]{3}{*}{\makecell[c]{Human\\Evaluation}}
& \cellcolor{gray!10} TA$\downarrow$
& \cellcolor{gray!10} 43.480
& \cellcolor{gray!10} 40.070
& \cellcolor{gray!10} 35.840
& \cellcolor{gray!10} 32.480
& \cellcolor{gray!10} \textbf{30.000}
& \cellcolor{gray!10} 50.000
& \cellcolor{gray!10} 41.670
& \cellcolor{gray!10} 36.130
& \cellcolor{gray!10} 34.890
& \cellcolor{gray!10} \textbf{33.060} \\

& TCons$\downarrow$
& 43.480
& 41.670
& 35.260
& 33.480
& \textbf{30.910}
& 50.000
& 43.670
& 37.610
& 33.330
& \textbf{32.010} \\

& \cellcolor{gray!10} MF$\downarrow$
& \cellcolor{gray!10} 48.630
& \cellcolor{gray!10} 42.940
& \cellcolor{gray!10} 37.560
& \cellcolor{gray!10} 35.270
& \cellcolor{gray!10} \textbf{31.640}
& \cellcolor{gray!10} 41.360
& \cellcolor{gray!10} 37.980
& \cellcolor{gray!10} 36.020
& \cellcolor{gray!10} 34.300
& \cellcolor{gray!10} \textbf{29.730} \\

\hline
\end{tabular}
}
\end{table}

\begin{figure*}[t]
\centering
\includegraphics[width=\textwidth]{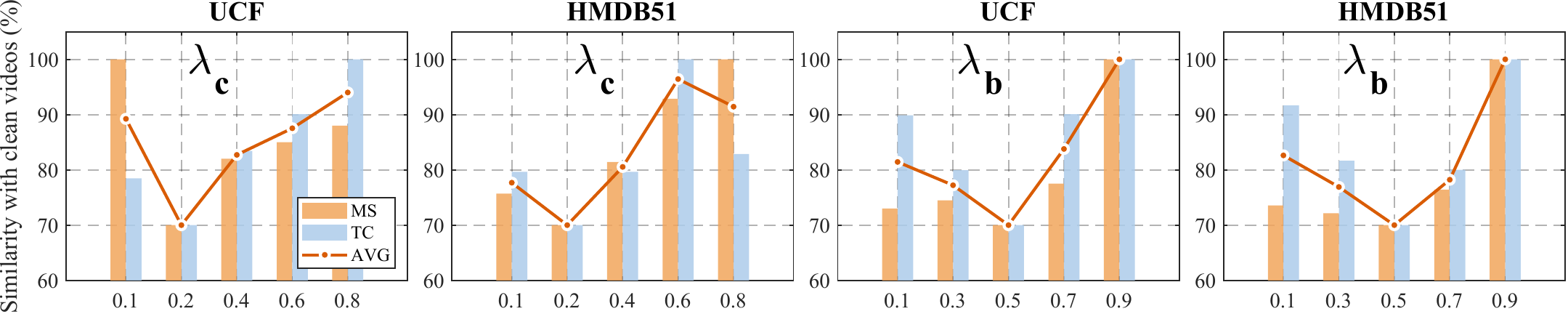} 
\vspace{-6mm}
\caption{Hyperparameter sensitivity analysis: Orange line is the average across two metrics.}
\label{fig:lambda_sensitivity}
\vspace{-2mm}

\end{figure*}

\begin{table}[H]
\centering
\footnotesize
\setlength{\tabcolsep}{4.0pt}
\renewcommand{\arraystretch}{1.08}
\caption{Ablation study of objective variants: automatic metrics and human preference scores.}
\label{tab:ablation_ucf_hmdb}
\resizebox{\textwidth}{!}{%
\begin{tabular}{l||cc|ccc||cc|ccc}
\hline
\rowcolor{gray!20}
\textbf{Dataset} $\rightarrow$
& \multicolumn{5}{c||}{\cellcolor{gray!20}\textbf{UCF}}
& \multicolumn{5}{c}{\cellcolor{gray!20}\textbf{HMDB51}} \\

\rowcolor{gray!20}
\textbf{Evaluation} $\rightarrow$
& \multicolumn{2}{c|}{\cellcolor{gray!20}\textbf{Automatic Evaluation}}
& \multicolumn{3}{c||}{\cellcolor{gray!20}\textbf{Human Evaluation}}
& \multicolumn{2}{c|}{\cellcolor{gray!20}\textbf{Automatic Evaluation}}
& \multicolumn{3}{c}{\cellcolor{gray!20}\textbf{Human Evaluation}} \\

\rowcolor{gray!20}
\textbf{Variant} $\downarrow$, \textbf{Metric} $\rightarrow$
& \cellcolor{gray!20}\textbf{MS}$\downarrow$
& \cellcolor{gray!20}\textbf{TC}$\downarrow$
& \cellcolor{gray!20}\textbf{TA}$\downarrow$
& \cellcolor{gray!20}\textbf{TCons}$\downarrow$
& \cellcolor{gray!20}\textbf{MF}$\downarrow$
& \cellcolor{gray!20}\textbf{MS}$\downarrow$
& \cellcolor{gray!20}\textbf{TC}$\downarrow$
& \cellcolor{gray!20}\textbf{TA}$\downarrow$
& \cellcolor{gray!20}\textbf{TCons}$\downarrow$
& \cellcolor{gray!20}\textbf{MF}$\downarrow$ \\
\hline\hline

Clean
& 0.995 & 4.662 & 70.000 & 69.090 & 68.360
& 0.995 & 4.538 & 66.940 & 67.990 & 70.270 \\

\cellcolor{gray!10}$\mathcal{L}_{\mathrm{fit}}+\mathcal{L}_{\mathrm{rel}}$
& \cellcolor{gray!10}0.990
& \cellcolor{gray!10}4.559
& \cellcolor{gray!10}47.500
& \cellcolor{gray!10}46.910
& \cellcolor{gray!10}44.720
& \cellcolor{gray!10}0.992
& \cellcolor{gray!10}4.492
& \cellcolor{gray!10}49.290
& \cellcolor{gray!10}48.480
& \cellcolor{gray!10}43.130 \\

$\mathcal{L}_{\mathrm{tem}}$
& 0.989 & 4.292 & 39.380 & 34.280 & 34.920
& 0.992 & 4.492 & 37.280 & 35.870 & 31.050 \\

\cellcolor{gray!10}$\mathcal{L}_{\mathrm{fit}}+\mathcal{L}_{\mathrm{rel}}+\mathcal{L}_{\mathrm{bdry}}$
& \cellcolor{gray!10}0.988
& \cellcolor{gray!10}4.264
& \cellcolor{gray!10}32.610
& \cellcolor{gray!10}40.710
& \cellcolor{gray!10}38.780
& \cellcolor{gray!10}0.989
& \cellcolor{gray!10}4.388
& \cellcolor{gray!10}35.670
& \cellcolor{gray!10}39.500
& \cellcolor{gray!10}35.270 \\
\hline

\rowcolor{cyan!12}
\textbf{\ChronoLock}
& \textbf{0.980} & \textbf{4.149} & \textbf{30.000} & \textbf{30.910} & \textbf{31.640}
& \textbf{0.988} & \textbf{4.333} & \textbf{33.060} & \textbf{32.010} & \textbf{29.730} \\
\hline
\end{tabular}
}
\end{table}

\vspace{-2mm}
\subsection{Additional Analysis}
\vspace{-2mm}

\textbf{Model Mismatch.}
The first transfer scenario is when the generation backbones are mismatched. We
provide examples of transferring perturbations trained on ZeroScope to defend
ModelScope personalization, and vice versa, as shown in
Table~\ref{tab:mismatch_results}. \ChronoLock still provides effective
protection under these cross-model settings. We also examine both transfer
directions across UCF and HMDB51, where the protected generations continue to
show degraded temporal quality and low human preference. These results suggest
that \ChronoLock does not merely overfit to one specific T2V backbone, but
learns perturbations that transfer across related video diffusion generators.

\textbf{Prompt Mismatch.}
The malicious user can also train the motion personalization model with a
different prompt from the one used during protection. \ChronoLock still degrades
motion smoothness, temporal consistency, and motion fidelity under this prompt
mismatch, as in Table~\ref{tab:mismatch_results}.

\textbf{Impact of Perturbation Budget.}
We evaluate the impact of varying perturbation budgets by testing
\ChronoLock under $\epsilon=0.01,.02,.03,.04$, and $.05$. As shown in
Table~\ref{tab:epsilon_sensitivity}, larger perturbation budgets generally lead
to stronger protection across both automatic and human evaluations. In
particular, $\epsilon=0.05$ achieves the best overall results across the two
datasets, with consistently stronger human-evaluation preferences and favorable
automatic scores. We use $\epsilon=0.05$ as the default perturbation
budget in our main experiments.

\textbf{Hyper-Parameter Analysis.}
In Fig.~\ref{fig:lambda_sensitivity}, we analyze the weights of the adjacent-frame variation term $\mathcal{L}_{\mathrm{var}}$ and the boundary disruption term $\mathcal{L}_{\mathrm{bdry}}$. The figure reports the similarity between protected-video and clean-video personalization under MS and TC, so lower values indicate stronger protection. Both datasets have the lowest average similarity at $\lambda_c\!=\!0.2$ and $\lambda_b\!=\!0.5$, suggesting that moderate weights best balance local temporal inconsistency and cross-chunk boundary disruption with the diffusion fitting and frame-relative objectives. Smaller weights provide insufficient temporal or boundary disruption, while overly large weights make the optimization less balanced and weaken the protection effect. Thus, we use $\lambda_c\!=\!0.2$ and $\lambda_b\!=\!0.5$ as the default setting in all main experiments.

\textbf{Ablation Study.}
The results are presented in Table~\ref{tab:ablation_ucf_hmdb}. Compared with
the clean videos, using only $\mathcal{L}_{\mathrm{fit}}+\mathcal{L}_{\mathrm{rel}}$
already reduces both automatic and human evaluation scores, showing that the
fitting and frame-relative relation terms provide a basic protection effect.
The variant with $\mathcal{L}_{\mathrm{tem}}$ further lowers TC and human
temporal-related scores, indicating that temporal disruption is important for
weakening personalized video generation. Adding $\mathcal{L}_{\mathrm{bdry}}$
further improves the overall protection effect, especially on MS, TC, and
human-evaluation metrics, confirming that boundary-aware optimization helps
disrupt cross-chunk motion continuity. However, this variant is still less
effective than the full method, particularly on TCons and MF, suggesting that
boundary guidance alone cannot fully produce the coordinated degradation
required for robust protection. Overall, \ChronoLock achieves the lowest MS,
TC, TA, TCons, and MF across both datasets. These consistent gains demonstrate
that the synergy among temporal regularization
and boundary-aware guidance is essential for effectively disrupting unauthorized personalization.

\vspace{-2mm}
\section{Conclusion}
\vspace{-2mm}
This paper presents \ChronoLock, a proactive protection framework tailored for safeguarding videos from unauthorized text-to-video personalization. Instead of directly extending image-level protective perturbations to videos, we target the temporal denoising dynamics that enable motion personalization. Through intra-chunk temporal disruption, \ChronoLock corrupts local motion adaptation by jointly enlarging diffusion fitting errors, frame-relative denoising mismatch, and adjacent-frame variation. Furthermore, by applying adaptive inter-chunk boundary disruption, it weakens long-range motion continuity across video segments. Extensive experiments on UCF Sports and HMDB51 with representative T2V backbones demonstrate that \ChronoLock consistently suppresses unauthorized motion imitation under both automatic metrics and human evaluation, while maintaining visually subtle perturbations.

\FloatBarrier
\bibliographystyle{unsrt}
\bibliography{ref}


\newpage
\appendix

\section{Details of Human Evaluation}
\label{HumanEvaluationDetails}

Following the human evaluation protocol used for generated videos in
MotionDirector~\citep{zhao2024motiondirector}, we conduct a pairwise human
evaluation through a browser-based annotation interface. For each evaluation
setting, we construct pairwise comparison tasks between the videos generated
from clean personalization and those generated from \ChronoLock-protected
personalization. The task pool covers all evaluated prompts and video examples.
During annotation, tasks are randomly sampled and assigned to raters, while we
enforce a coverage constraint to ensure that every example in the task pool is
evaluated at least once. Each comparison task is completed by four human raters,
none of whom participated in the experiments or had access to the identities of
the compared methods.

\begin{figure}[H]
\centering
\includegraphics[width=0.94\textwidth,trim=130pt 85pt 130pt 105pt,clip]{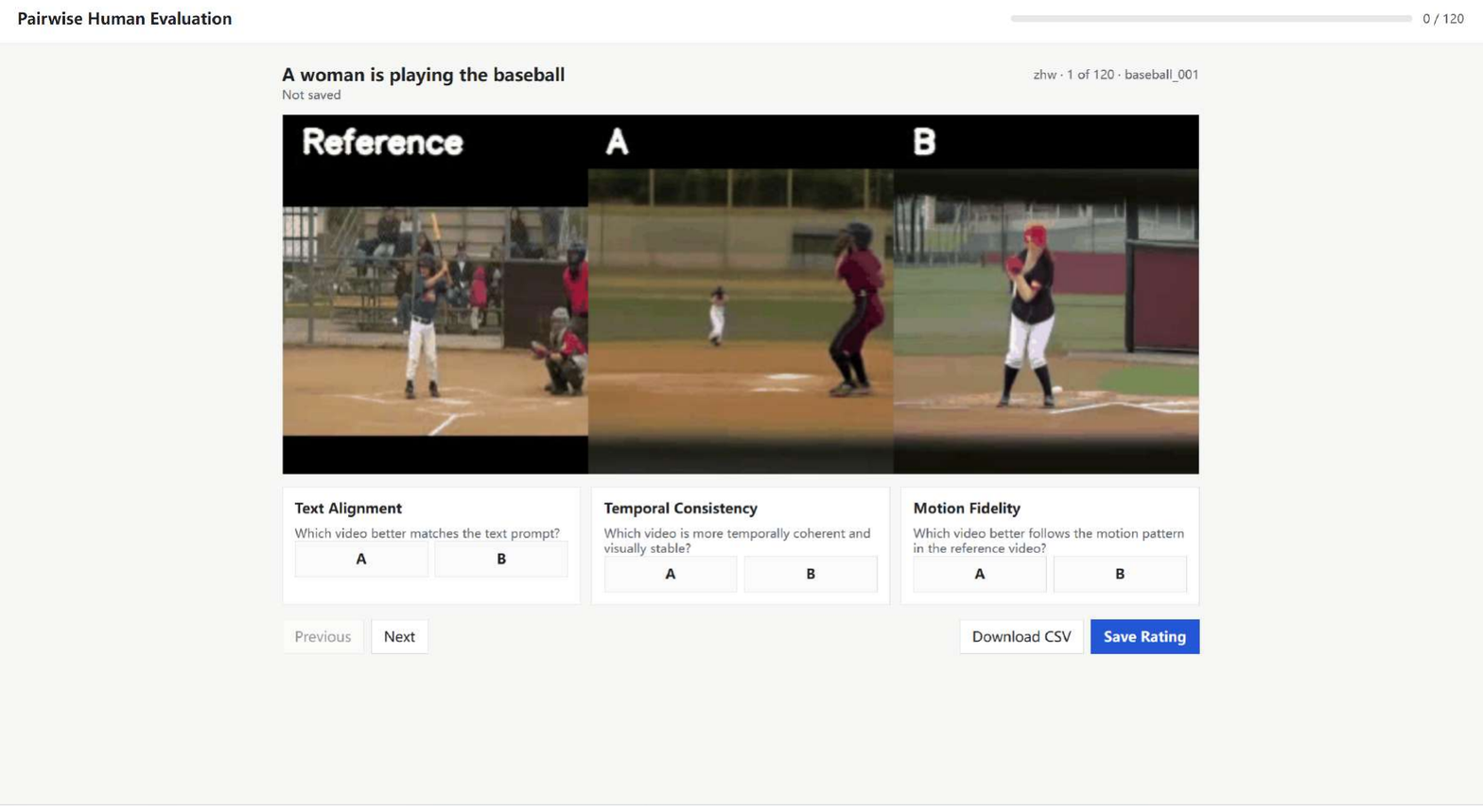}
\caption{Browser-based interface for pairwise human evaluation. Each task shows
the text prompt, a reference video, and two anonymized generated videos assigned
to options A and B. Raters answer three questions covering text alignment,
temporal consistency, and motion fidelity.}
\label{fig:human_eval_interface}
\end{figure}

To avoid positional bias, the two compared methods are randomly mapped to
options A and B for each task. As shown in Fig.~\ref{fig:human_eval_interface},
each task presents the text prompt at the top of the page, followed by three
videos arranged side by side: the reference video, the video assigned to option
A, and the video assigned to option B. Raters answer the following three
questions.

\textbf{Text Alignment.} Which video better matches the text prompt?

\textbf{Temporal Consistency.} Which video is more temporally coherent and
visually stable?

\textbf{Motion Fidelity.} Which video better follows the motion pattern in the
reference video?

For the motion fidelity question, raters are instructed to compare the motion
patterns against the reference video rather than focusing on appearance or
style. The final human-evaluation scores are computed as the preference
percentage of each method over all completed pairwise comparisons.

\section{Limitations}
\label{app:limitations}

\ChronoLock is designed for proactive protection against unauthorized
text-to-video personalization, with experiments instantiated on MotionDirector
Temporal LoRA. This threat model matches settings where an attacker adapts
temporal modules from released reference videos, but it does not cover every
possible misuse of video data. For example, full-model fine-tuning, future
personalization pipelines, or attacks that combine video protection removal with
manual curation may require additional evaluation.

The current evaluation focuses on HMDB51 and UCF Sports Actions, with ZeroScope
and ModelScope as generation backbones. These datasets and models cover common
human-action videos and representative open T2V backbones, but they do not
exhaust all domains where video personalization may be used. Larger in-the-wild
video collections, higher-resolution long videos, and other motion categories
remain important settings for future study.

Finally, \ChronoLock is evaluated against common transformations and
cross-model transfer settings, but it does not provide a formal guarantee
against all adaptive preprocessing or restoration attacks. The protected videos
are constrained by an \(\ell_\infty\) budget and evaluated with distortion
metrics, but some applications may require stricter visual-quality or
downstream-utility constraints before deployment.

\section{Broader Impact}
\label{app:impact}

This work studies protection for videos that may otherwise be collected and used
for unauthorized T2V personalization. A positive impact is that video owners may
gain a technical tool for reducing unwanted imitation of their motion patterns,
styles, or visual identities when videos are shared online. This is relevant to
creative media, personal videos, sports clips, and other settings where video
data can be reused without consent.

From a research perspective, \ChronoLock highlights that video data protection
should consider temporal denoising dynamics rather than only frame-level
appearance cues. We hope this encourages more careful evaluation of temporal
misuse risks in generative video models and more transparent discussion of data
ownership in personalization systems.

The method also has potential negative uses. Protective perturbations could be
applied to public or shared videos to hinder legitimate training, benchmarking,
or reproducibility. \ChronoLock should not be treated as a substitute for
consent, licensing, access control, or platform-level policy. Responsible use
requires that the protector owns or is authorized to modify and release the
videos being protected.

\section{Compute Resources}
\label{app:compute}

All experiments were conducted on four NVIDIA RTX A6000 GPUs. The computation mainly
consists of generating protected videos with the MotionDirector Temporal LoRA
surrogate and evaluating downstream T2V personalization on the selected
backbones and datasets. The total compute increases with the number of datasets,
generation backbones, perturbation settings, robustness tests, ablations, and
cross-model transfer experiments.

\section{Additional Results}
\label{app:additional_result}

\begin{figure}[H]
\centering
\includegraphics[width=\textwidth]{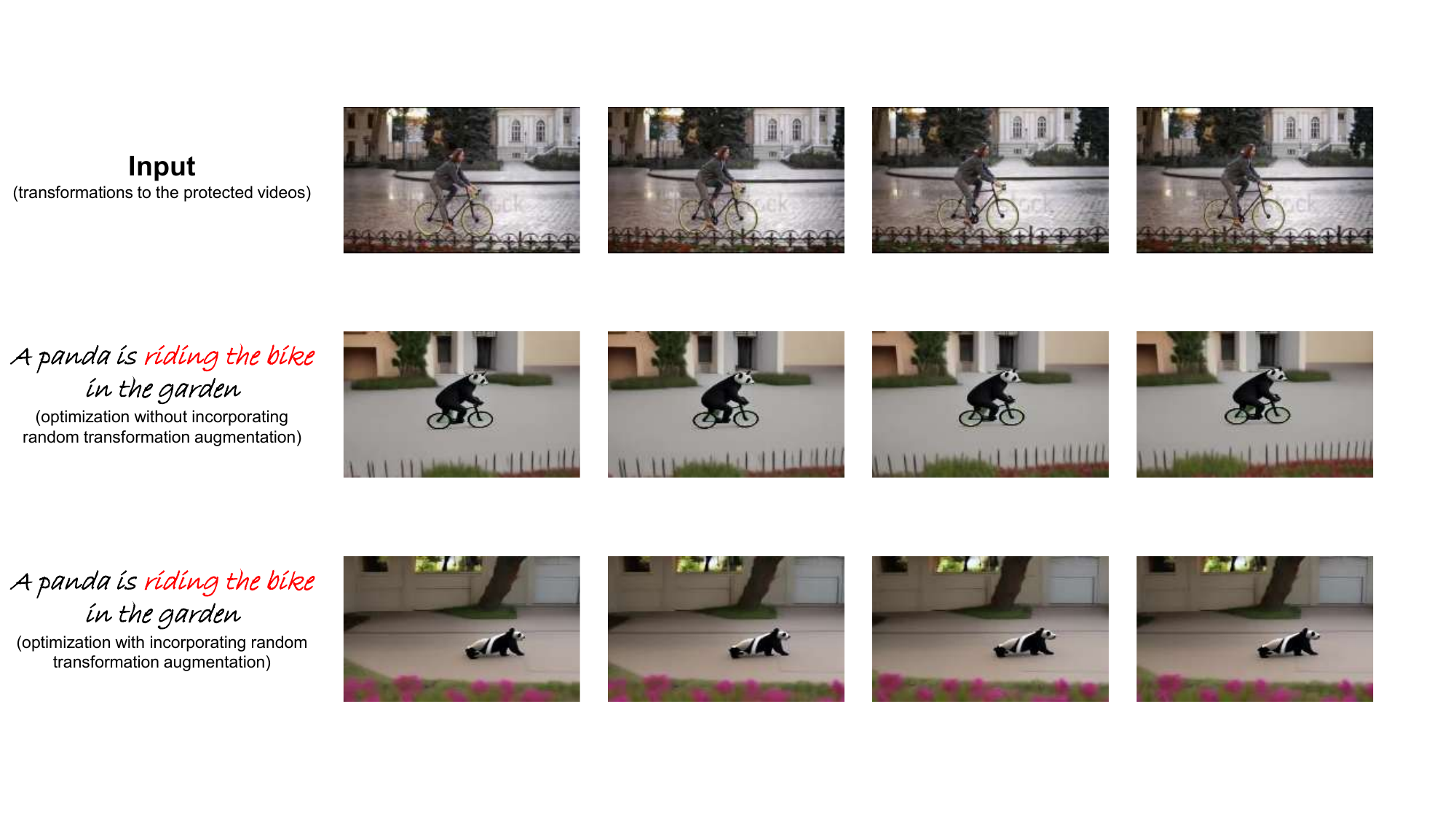}
\caption{Additional qualitative results on transformation-sampled optimization.
Compared with optimization without random transformation augmentation,
\ChronoLock with transformation sampling more strongly disrupts unauthorized
motion personalization.}
\label{fig:add_transform}
\end{figure}

We provide additional qualitative results to further illustrate the behavior of
\ChronoLock under different protection settings. Fig.~\ref{fig:add_transform}
shows the effect of transformation-sampled optimization. When random
transformation augmentation is not used, the personalized model can still
recover a temporally plausible motion pattern from the protected reference
video. In contrast, incorporating random transformations during perturbation
optimization leads to stronger degradation after unauthorized personalization:
the generated subject fails to preserve the target motion and exhibits weaker
temporal continuity.

\begin{figure}[H]
\centering
\includegraphics[width=\textwidth]{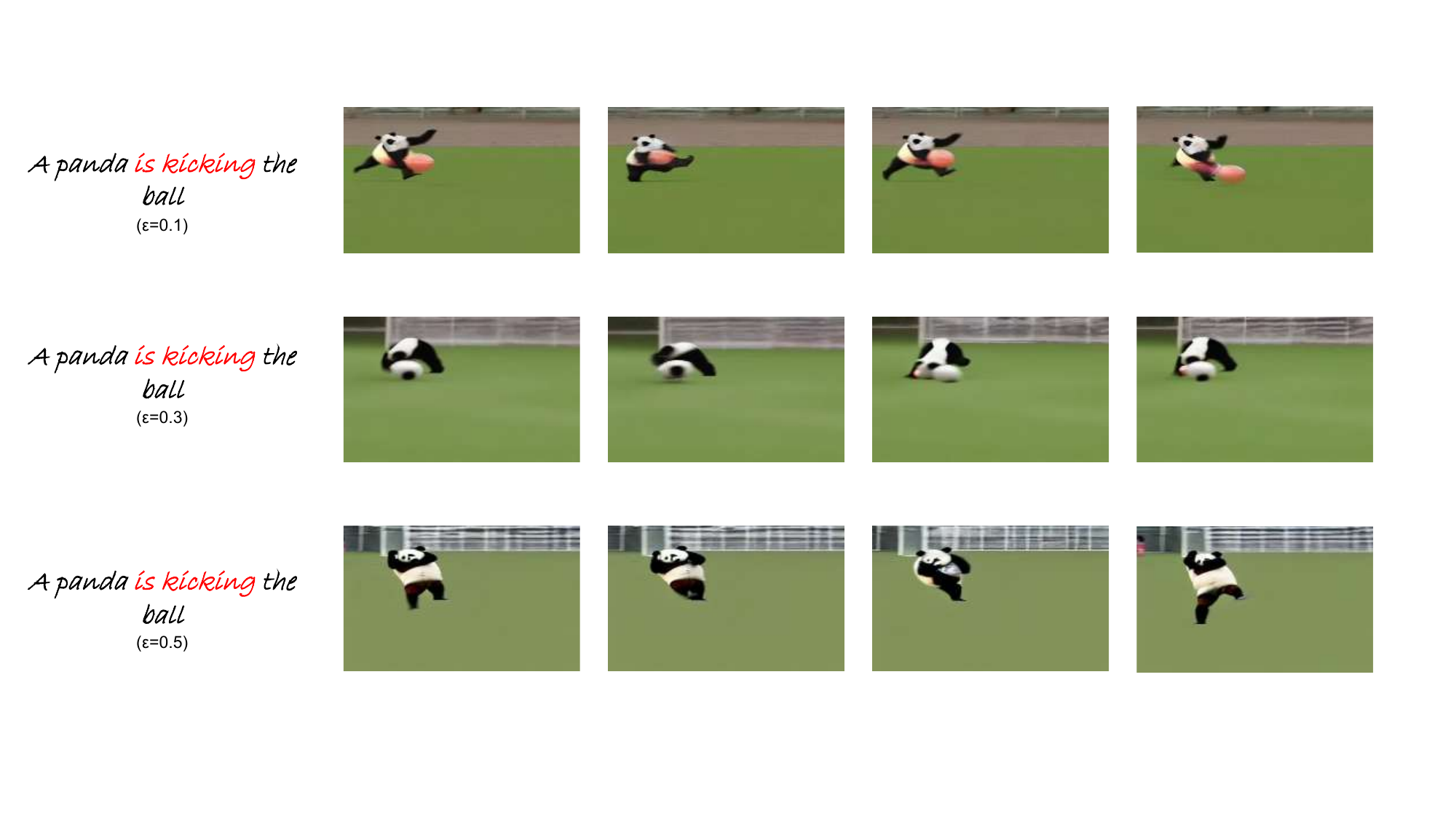}
\caption{Additional qualitative results under different perturbation budgets
$\varepsilon$. Larger budgets lead to stronger disruption of motion fidelity and
temporal consistency after unauthorized personalization.}
\label{fig:add_epsilon}
\end{figure}

Fig.~\ref{fig:add_epsilon} visualizes the effect of the perturbation budget
$\varepsilon$. As $\varepsilon$ increases, the generated videos become less
faithful to the reference motion, showing that a larger perturbation budget
provides stronger protection against motion customization. This qualitative
trend is consistent with the quantitative results in
Table~\ref{tab:epsilon_sensitivity}.

\begin{figure}[H]
\centering
\includegraphics[width=\textwidth]{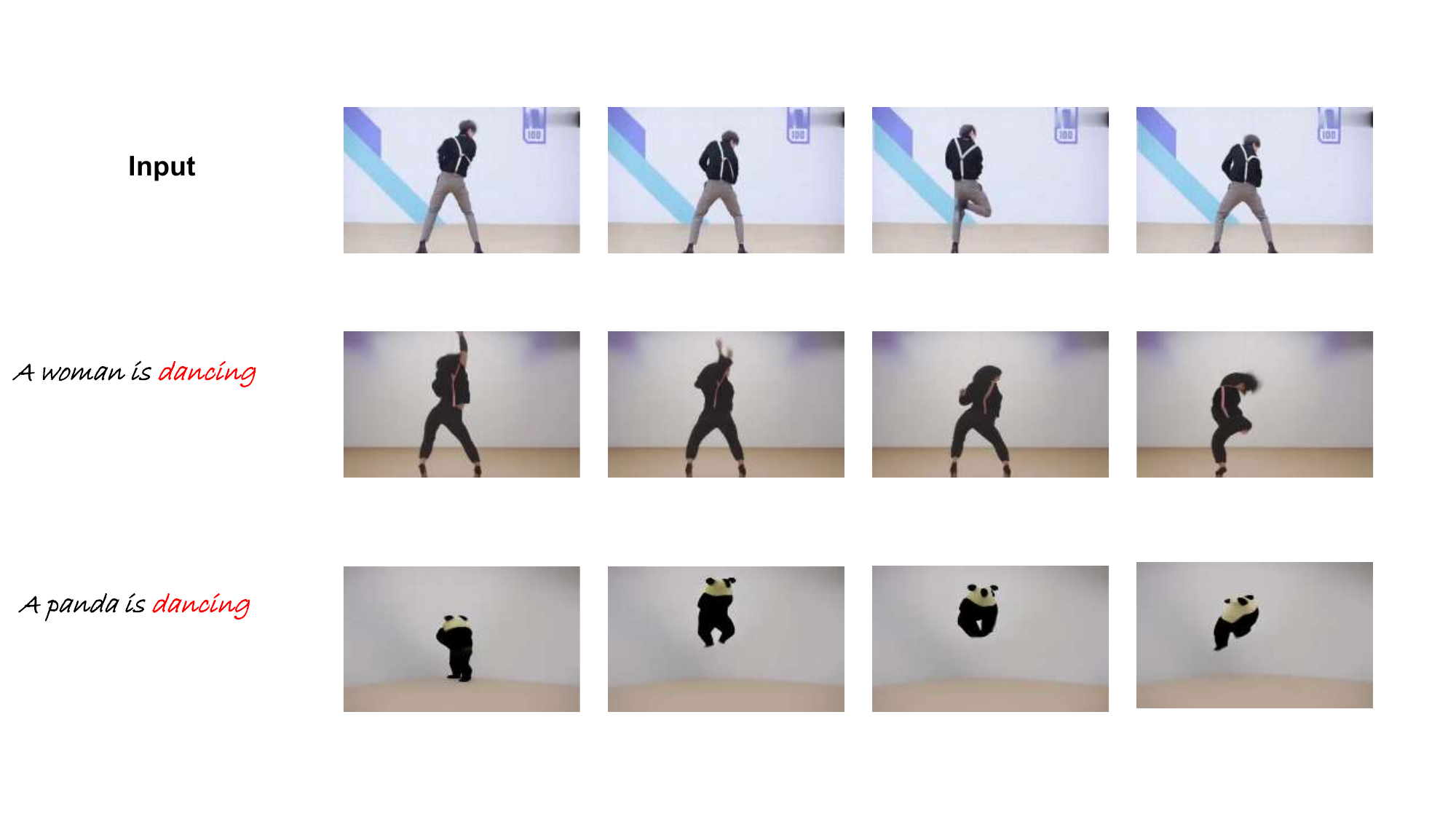}
\caption{Additional prompt-level qualitative results. \ChronoLock disrupts
motion personalization under different target prompts, indicating that the
protection affects temporal motion learning rather than only static appearance.}
\label{fig:add_prompt_examples}
\end{figure}

We also show additional prompt-level examples in
Fig.~\ref{fig:add_prompt_examples}. Given the same protected reference video,
unauthorized personalization is evaluated with different prompts, such as
``a woman is dancing'' and ``a panda is dancing''. The results show that
\ChronoLock disrupts the transferred motion pattern even when the generated
appearance or text prompt changes, suggesting that the protection targets the
temporal personalization signal rather than only static visual appearance.

\begin{figure}[H]
\centering
\includegraphics[width=\textwidth]{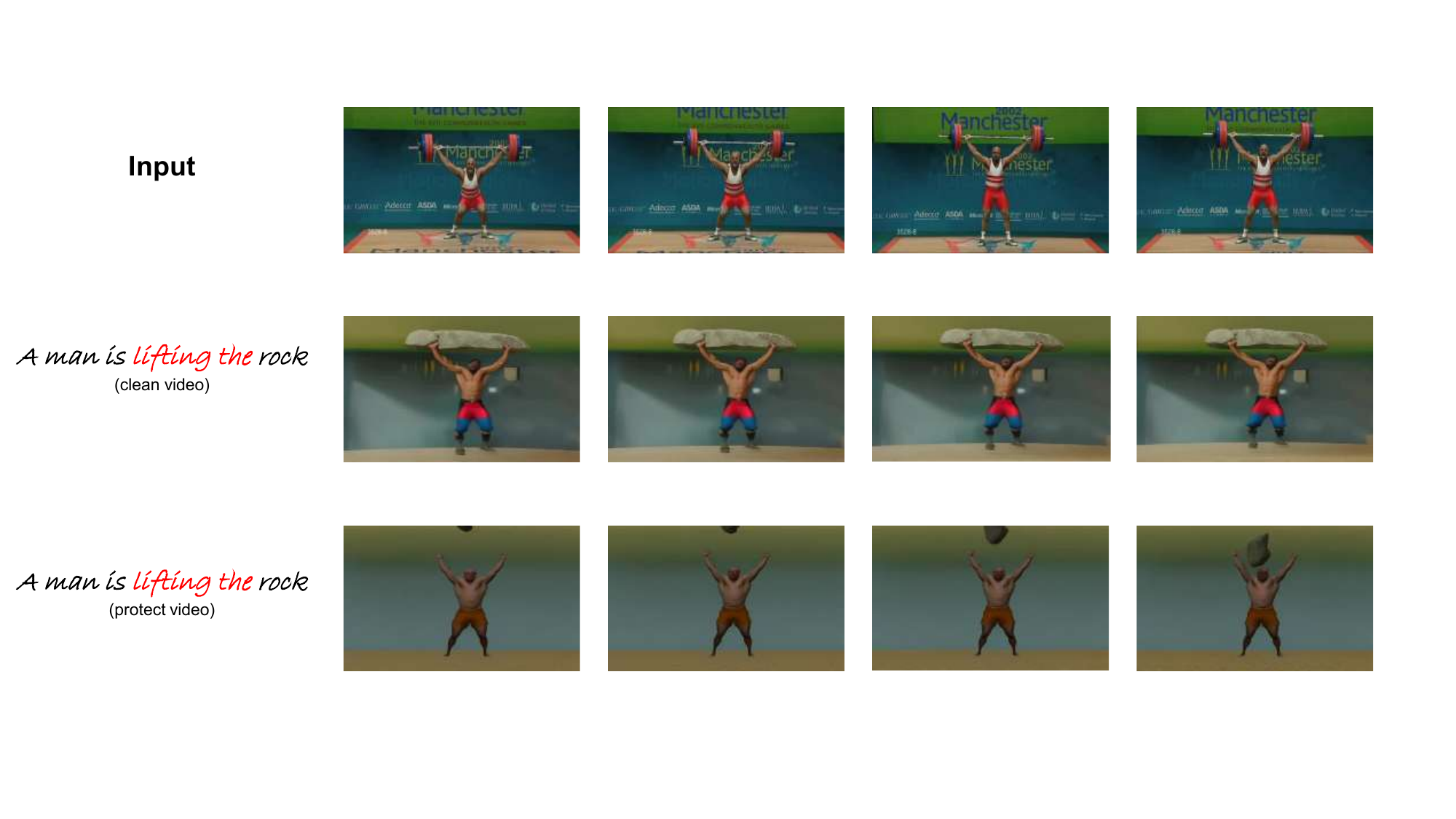}
\caption{Comparison between clean-video personalization and
\ChronoLock-protected personalization. Protected videos lead to less faithful
motion imitation and weaker temporal coherence in the customized generations.}
\label{fig:add_clean_protected}
\end{figure}

Finally, Fig.~\ref{fig:add_clean_protected} compares personalization results
from clean videos and \ChronoLock-protected videos. The model trained on clean
videos preserves the reference action with coherent temporal progression,
whereas the model trained on protected videos produces less faithful motion and
weaker frame-to-frame consistency. These examples further support that
\ChronoLock effectively suppresses unauthorized motion imitation while keeping
the released reference videos visually close to the originals.

\end{document}